\documentclass[journal]{IEEEtran}
\usepackage[pdftex]{graphicx}
\usepackage{algorithm}
\usepackage[noend]{algpseudocode}
\usepackage{amsthm}

\makeatletter
\renewcommand{\Function}[2]{%
  \csname ALG@cmd@\ALG@L @Function\endcsname{#1}{#2}%
  \def\jayden@currentfunction{#1}%
}
\newcommand{\funclabel}[1]{%
  \@bsphack
  \protected@write\@auxout{}{%
    \string\newlabel{#1}{{\jayden@currentfunction}{\thepage}}%
  }%
  \@esphack
}
\makeatother

\ifCLASSINFOpdf
\else
\fi
\begin{document}

\title{Mining useful Macro-actions in Planning}

 \author{Sandra Castellanos-Paez$^{\textbf{*}}$, Damien Pellier$^{*}$, Humbert Fiorino$^{*}$ and Sylvie Pesty$^{*}$
 \thanks{$^{*}$ Univ. Grenoble Alpes, LIG, F-38000 Grenoble, France
     \texttt{firstname.lastname at imag.fr}}%
 }

%



\maketitle

\begin{abstract}
Planning has achieved significant progress in recent years. Among the various approaches to scale up plan synthesis, the use of macro-actions has been widely explored.
As a first stage towards the development of a solution to learn on-line macro-actions, we propose an algorithm to identify useful macro-actions based on data mining techniques.
The integration in the planning search of these learned macro-actions shows significant improvements over six classical planning benchmarks.
\end{abstract}


%
\IEEEpeerreviewmaketitle

\section{Introduction}

Automated planning is an area of Artificial Intelligence that comes up with the challenge of devising systems that can autonomously find a plan to reach a set of goals. In classical planning, a problem is composed of an initial state, a goal specification and a set of actions. From the initial state if the preconditions of an action are satisfied, the action is applicable to the current state. Thereby, the action effects can be applied to generate a new state. This is done for each new state until the goal is reached. The solution of a planning problem is a plan. A plan defines a sequence of actions from the initial state to the goal state for a given problem over a domain.

Planning  is NP-hard. Thus, the focus remains on developing powerful planning techniques capable of effectively explore the search space that grows exponentially. A way to increase planner performance consists in exploiting knowledge about the structure of the planning tasks \cite{IPC:2002}. This could be accomplished by using {\it macro-actions}, i.e., a sequence of actions that occurs frequently in solution plans. Macro-actions have been widely studied among the different approaches to speed-up planning processes. Learning macro-actions from previously acquired knowledge (plans) allows to go deep quickly into the search space by triggering them during the search.

\begin{figure}
\begin{center}
\includegraphics[scale=0.5]{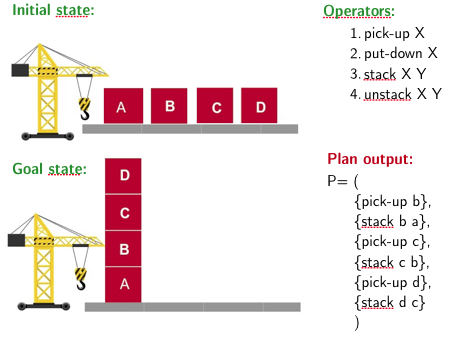}
\vspace{-1em}
\caption{Blocksworld domain.}
\vspace{-2em}
\label{Fig:bw}
\end{center}
\end{figure}

The use of macro-actions is not arbitrary. If a sequence of actions has a higher frequency on different plans, reasoning lead us to consider it as a candidate of useful sequences for a given domain.
Let's have an example, consider the \textit{blocksworld} planning domain in Figure \ref{Fig:bw}. The goal is to stack a set of blocks. This domain has five operators: pick-up, picks a block $x$ from the table; put-down, puts a block $x$ on the table; stack, puts a block $x$ on a block $y$; unstack, removes a block $x$ from a block $y$. It is logical to suggest that once we pick a block from the table the next most probably action will be stack it on another block or put it down. If we learn one of these sequences as a macro-action e.g. pick-up\_stack, we can apply it directly avoiding the analysis of what action comes after the pick-up action.

\begin{figure}
\begin{center}
\includegraphics[scale=0.5]{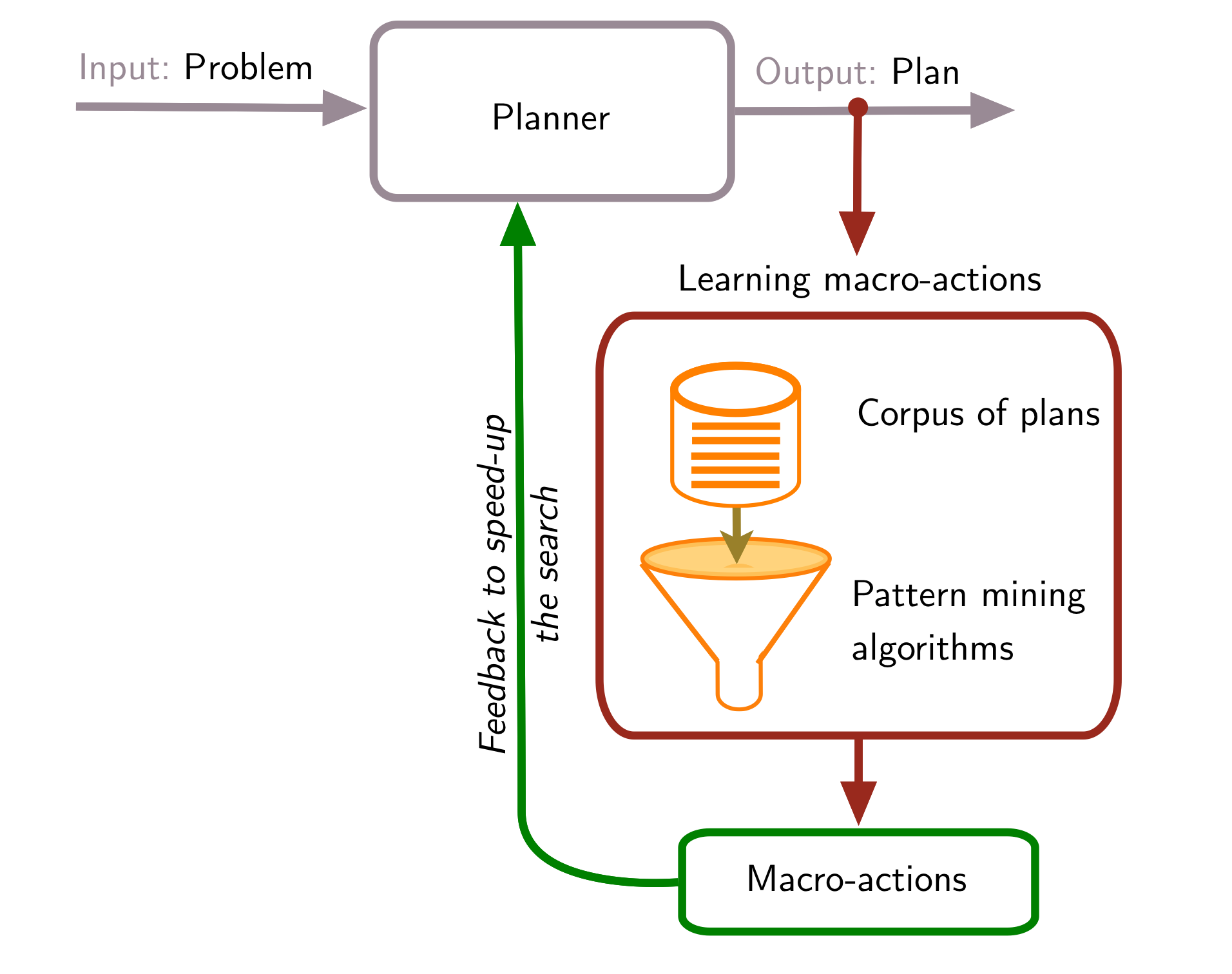}
\vspace{-1em}
\caption{Mining Framework.}
\vspace{-1em}
\label{Fig:framework}
\end{center}
\end{figure}

\begin{figure}
\begin{center}
\includegraphics[scale=0.5]{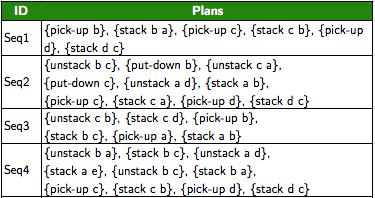}
\vspace{-1em}
\caption{Corpus of plans.}
\vspace{-2em}
\label{Fig:plans}
\end{center}
\end{figure}

Figure \ref{Fig:framework} shows our framework to learn useful sequences as macro-actions and use them to speed-up the search. We solve some problems and we keep the output plans in order to build a sequence database (Figure \ref{Fig:plans}). Then, we look for useful sequences by using a sequential pattern mining algorithm. This algorithm finds the complete set of patterns satisfying a minimum frequency threshold. Every frequent sequence of that set represents a macro-action. Finally, we feed the planner with the macros to save time searching.

Although much work has been done to date \cite{Botea:macro-ff,Newton:2007,Jonsson:2009,Dulac:2013, Chrpa:2014}, more studies need to be conducted to ensure successful results in the performance of planning engines in exploiting macros regardless of the planning domain. The purpose of this research was to further investigate a strategy for detecting useful macros based in pattern mining algorithms. We propose a novel planner independent macro learning method. First, to detect automatically useful macro-actions trough sequential pattern mining algorithms and second, to implement these macro-actions in any state-space search algorithm to speed-up the search.

The paper is organized as follows. We will first introduce a literature review in macro-actions past works. Then we will present the planning concepts used in this work followed by the motivation for using pattern mining algorithms. After that, we will get into the core of the proposed method for generating and using macro-actions. Finally, we will discuss the results and the possible future work.

\section{Related Work}
We group macro-actions related work into two main categories: off-line and on-line techniques.\label{RW} An off-line approach offers as an advantage an ease view over the macro-actions use, but also over the impact in the search time. Into this category, we found the adaptive planning system Macro-FF \cite{Botea:macro-ff}. It extracts macro-actions from solutions of training problems, and by identifying statically connected abstract components. Only the macro-actions showing effective performances in solving training problems are kept for future searches. Newton et al. \cite{Newton:2007} proposed another offline method which uses a genetic algorithm as a learning technique and plans as the macro generation source. The algorithm generates the macros from plans of simple problems to seed the population and evaluates them through a ranking method based on the weighted average of time differences in solving more difficult problems with the original domain augmented with macros. In a more recent work, Dulac et al. \cite{Dulac:2013} introduced a domain independent approach for learning macros from before computed solutions. It extracts statistical information from successful plans based on a n-gram analysis. Then it builds a macro library based on earlier information, a generalisation and a specialisation process. Finally, it adds selected macros into the planning domain after a filtering phase based on statistical information and heuristics. Later, Chrpa et al. proposed a technique \cite{Chrpa:2014} to maximise the utility of macros. It first learns the causal relations between planning operators and initial or goal predicates (also known as outer entanglements) by using an approximation algorithm in several training plans. Then, exploiting this knowledge it extracts macros and uses them to reformulate the original domain model.

By contrast, an on-line approach remove the need of extra training problems and off-line filtering. Coles and Smith described Marvin planner in \cite{Coles:2007a}. It identifies regions in the search space where the heuristic values of all successors is greater than or equal to the best seen so far. Then, it learns the escaping macro-actions to use them in similar regions during the search. This work was improved in \cite{Coles:2007} by keeping libraries of macro-actions for use on future problems. Chrpa et al.\cite{Chrpa:2015} extended their early technique by generating useful macros from outer entanglements in the search without an offline learning phase.

Other approaches presented an algorithm \cite{Jonsson:2009} to decompose a domain in several subproblems to then solve them. After, the algorithm stores each generated partial plan in memory as a macro which allows to retrieve and use it as part of the solution for a different problem that contains it. Masataro and Fukunaga introduced a similar work in \cite{Masataro:2015} which automatically identifies subproblems, generate macros from subplans and integrate the subplans by solving the augmented problem.

In spite of macro-actions can be intuitively built from sequence of actions occurring many times in solution plans, there is no previous work based on mining frequent sequences through data mining techniques. Also, no previous research on using macro-actions in automated planning has fully implemented a planner independent approach.

\section{Pattern mining for macro learning}

\subsection{Planning system definition}
\begin{sloppypar}

We address sequential planning in the STRIPS framework \cite{finkes:71}. An {\it action} $a$ is a tuple $a = (pre(a), add(a), del(a))$ where $pre(a)$ are the action's {\it preconditions}, $add(a)$ and $del(a)$ are respectively its positive and negative {\it effects}.

A {\it state} $s$ is a set of logical propositions. A state $s'$ is reached from $s$ by applying an action $a$ according to the transition function $$\gamma(s,a) = (s - del(a)) \cup add(a).$$ The application of a sequence of actions $\pi = \langle a_1, \ldots, a_n \rangle$ to a state $s$ is recursively defined as $\gamma(s,  \langle a_1, \ldots, a_n \rangle) = \gamma(\gamma(s, a_1), \langle a_2, \ldots, a_{n} \rangle)$.

A {\it planning domain} is composed of a set of actions and a set of predicates i.e. properties of objects. A {\it planning problem} is defined by a planning domain, an initial state and a set of goal states. Thus, a \textit{plan} is a sequence of actions $\pi = \langle a_1, \ldots, a_n \rangle$ such that the goal $g \subseteq \gamma(s,\pi)$ and $g$ is reachable if such a plan exists.

Planning problems have been shown PSPACE complete, it means the size of the search space is huge. Thus planning systems must reduce the size of the search space they traverse.

Classical approaches get as output a sequence of actions (plan) for every input problem from an input domain. However, for every new problem they execute the same process without keeping memory of past events. Real world problems give us an idea of encapsulating routines in few steps that have subroutines inside. For example, in everyday life if we are in a room and we need go to the kitchen a possible solution is $(open-door-room, go-out, walk, open-door-kitchen)$. In the same context if we need go to the bathroom a possible solution is $(open-door-room, go-out, walk, open-door-bathroom)$. By examining common sequences of actions when solving problems in a given domain, we can generalize a routine. In our example, we could rewrite the solutions as $go-from-roomA-to-roomB$.

\end{sloppypar}

\subsection{Interest towards pattern mining use}
\begin{sloppypar}
The discovery of recurring relationships among huge amount of data can help in the prediction of the next element in a sequence. \textit{Frequent patterns} are patterns that appear frequently in a data set. A \textit{sequence database} is a set of sequences where each sequence is a list of itemsets. A \textit{sequential pattern} is a sequence $s_a = x_1, x_2, \ldots , x_k$ (where $x_1, x_2, \ldots , x_k$ are itemsets) is said to occur in another sequence $s_b = y_1, y_2, \ldots , y_m$ (where $y_1, y_2, \ldots , y_k$ are itemsets) if and only if there exists integers $1 \leq i 1<i_2< \ldots <i_k \leq m$ such that $x_1 \subseteq y_{i1}, x_2\subseteq y_{i2}, \ldots, x_k \subseteq y_{ik}$. The \textit{support} of a sequential pattern is the number of sequences where the pattern occurs divided by the total number of sequences in the database.

A sequential pattern may have several forms: a \textit{frequent sequential pattern} is a sequential pattern having a support no less than the parameter provided by the user; a \textit{closed sequential pattern} is a sequential pattern that is not included in another pattern having the same support; a \textit{maximal sequential pattern} is a sequential pattern that is not strictly included in another closed sequential pattern. We are interested in the latter because the set is much smaller than the others.

Table \ref{Table:pmalgos} presents the studied algorithms of sequential pattern mining for this work. Besides sequential rule mining algorithms were also analyzed but discarded because we are strictly focused in ordered sequences.

\begin{table}[h!]
\centering
\begin{tabular}{|l||c|c|c|}
 \hline
 Algorithm & Form & Gap Config. & Max Length\\
 \hline
 CM-SPADE   & all    &no&   no\\
 CM-SPAM   & all    &yes&   yes\\
 BIDE+   & closed    &no&   no\\
 VMSP   & maximal    &yes&   yes\\
 MaxSP   & maximal    &no&   no\\
 \hline
\end{tabular}
\vspace{1em}
\caption{Sequential pattern mining algorithms}
\vspace{-2em}
\label{Table:pmalgos}
\end{table}

Detecting sequences of actions in a set of plans can be compared with mining sequential patterns in a data set. However, our interest remains in finding optimal sets of ordered sequences of actions where each consecutive action of a pattern must appear consecutively in a sequence i.e. no gap is allowed. Only the algorithms allowing the gap configuration interest us. Otherwise if the gap configuration is not allowed then the algorithms could find frequent sequences where actions are not consecutive. With regard to this, we focused in VMSP\footnote{Vertical mining of Maximal Sequential Patterns} algorithm \cite{fournier2014vmsp}.

VMSP mines a compact representation of the set of sequential patterns making it easier for users to analyze results. It uses a depth-first exploration of the search space using a vertical representation. First, it implements a search procedure to construct the set of frequent sequences $F_1$ given a \textit{minsup} threshold. Then, it filters the non maximal patterns from $F_1$ by checking from each sequence $s_a$ if there exists a pattern $s_b$ such that $s_a \sqsubseteq s_b$. Finally, it does a candidate pruning by exploiting item co-occurrence information.

\end{sloppypar}

\subsection{Implementation}
\begin{sloppypar}
The main idea of our approach is to build macro-actions from sequential patterns of actions in a set of plans and to use them during the planning search to improve the performance of the planning system.

A \textit{sequential pattern of actions} is a frequent action subsequence ($\alpha$) existing in a single plan or a set of plans ($\theta$). The support of a sequence $supp^\theta (\alpha)$ is the number of plans in $\theta$ that contains $\alpha$. Given a support threshold $\sigma$, a sequence $\alpha$ is a \textit{frequent sequence} on $\theta$ if $supp^\theta (\alpha) \geq \sigma$. Mining of maximal patterns consists of finding the set of \textit{maximal sequences} i.e. for a sequence $\alpha$ there is no other sequence $\beta$ such that $\alpha \sqsubseteq \beta$.


Macros should appear many times in solutions plans and encapsulate knowledge that is reusable across the problems in a domain. The first condition can be accomplished by studying the support parameter. If a sequence of actions occurs many times in plans, it might be a good sequence to examine. On the other hand, the second condition can take advantage of maximal sequences. A large number of sequences of actions makes it difficult to analyze results, then mining maximal sequences leads to a more compact set but also better efficiency.

Our method to learn and to use macro actions includes a \textit{set-up algorithm} and a \textit{enhanced search algorithm}. Algorithm \ref{alg:1} takes as input a learning set of non-empty solution plans. In line ~\ref{func:get-line}, it obtains a set of maximal sequential patterns by using the VMSP\cite{fournier2014vmsp} algorithm. Afterwards, from a encoded problem $P$ the function \ref{func:encode} evaluates each sequence from $R$. It checks if each one of the actions from the evaluated sequence belongs to the encoded operators of the problem. When the condition is met the whole sequence is encoded and added in the problem macro-action list (line ~\ref{var:table-line}).

\renewcommand{\algorithmicforall}{\textbf{for each}}
\begin{algorithm}[H]
\caption{Set-up algorithm}\label{algo1}
\begin{algorithmic}[1]
\State D $\gets$ Set of non-empty solution plans
\State A $\gets$ Algorithm for maximal sequence mining
\State R $\gets$ \textbf{SPMF}(D, A, supp)
\label{func:get-line}
\Function{encodeMacros}{encodedProblem P} \funclabel{func:encode}\label{func:encode-line}
  \ForAll {action c in GetCandidate(R)}
  \If {c $\in$ GetOperators (P)}
  \State T[indexOf(R)] $\gets$ Add ($O_i$)
  \label{var:table-line}
  \EndIf
  \EndFor
  \State \Return T
\EndFunction
\end{algorithmic}
\label{alg:1}
\end{algorithm}

The Algorithm \ref{alg:2},  takes as input the encoded macro-actions list obtained in Algorithm \ref{alg:1}. This algorithm can be implemented to any search algorithm in trees to speed-up the search. The purpose of this work is to validate the added value of this generic method. For this work, we chose a classical implementation of the A* algorithm. Line ~\ref{var:open-line} represents a priority queue while lines ~\ref{var:closeS-line} and ~\ref{var:openS-line} represent respectively the explored and pending nodes. In line ~\ref{var:curr-line}, a node $x$ is selected from the pending nodes list taking into account its heuristic value $h$. A plan is reached when $x$  satisfy the goal. If not the applicability of each  macro-action $m_i = \langle a_1a_2...a_n\rangle$  over $x$ is evaluated in line ~\ref{cond:app-line}. A macro-action is applicable in $x$  when the preconditions of $x$ satisfies the preconditions of $m_i[0]$ allowing to get the successor $x'$ and for each obtained successor  the next actions are applicable ( from $n>0, m_i[n-1]$ is applicable to $x^{n-1}$ ). The created successors update the list of pending nodes and the list of explored nodes (lines ~\ref{cond:update-line} and ~\ref{cond:remove-line}). After the algorithm tries to apply the problem operators. Finally, the node $x$ is added to the explored nodes and another node is selected from the pending nodes list.

\algrenewcommand\algorithmicindent{1.0em}
\renewcommand{\algorithmicforall}{\textbf{for each}}
\renewcommand{\algorithmicrequire}{\textbf{Input}}
\renewcommand{\algorithmicensure}{\textbf{Output}}
\begin{algorithm}[H]
\caption{Enhanced search algorithm}\label{algo2}
\begin{algorithmic}[1]
\Require macro-actions list $T [m_i <a_1,...,a_n>]$
\State P $\gets \{\}$
\State open $\gets$ root
\label{var:open-line}
\State closeset $\gets \{\}$
\label{var:closeS-line}
\State openset $\gets$ init
\label{var:openS-line}
\While{open $\neq$ null}
\State{current $\gets$ poll(open)}
\label{var:curr-line}
\If{current satisfy G} \State{extract (P)}
\Else
\ForAll {macro $m_i[a_1,a_2,...,a_n]$ in T}
  \If {$m_i$ isApplicableTo current}
  \label{cond:app-line}
  \ForAll {$s_i$ in generateStates(current, $m_i$)}
  \If {$s_i \in$ openSet and cost(s) $<$ cost ($s_i$)}
  \label{cond:update-line}
  \State{update (s, $s_i$)}
  \Else
  \If {$s_i\in$ closeSet and cost (s)$<$ cost ($s_i$)}
  \label{cond:remove-line}
  \State{remove (closeSet, $s_i$)}
  \EndIf
  \State{add (openSet, open, $s_i$)}
  \EndIf
  \EndFor
  \EndIf
  \EndFor
  \State{ applyOperators (current)}
\EndIf
\EndWhile
\Ensure P
\end{algorithmic}
\label{alg:2}
\end{algorithm}

\end{sloppypar}

\section{Results and Discussion}
\subsection{Experimental setup}
The experiments were based on barman, blocksworld, depots, ferry, grid and sokoban benchmarks. They were carried out on an Intel Xeon E5-2630 2.30GHz. The allocated CPU time was set to 300 seconds with a maximum of 8GB of memory. For each benchmark, a learning set of plans of 1000 problems and a test set of 300 problems were randomly generated with the generators\footnote{https://bitbucket.org/planning-tools/pddl-generators} used for the International Planning Competition (IPC).  We  went trough the SPMF \cite{JMLR:v15:fournierviger14a} data mining library, which implements the VMSP algorithm, to get the set of maximal sequences. We did not allow gaps and we varied the degree of support in steps of 1 from 1\% until no sequences were founded. We used the PDDL4J \footnote {https://github.com/pellierd/pddl4j} library to encode the sequences into a forward chaining planner based on A* algorithm and on FF heuristic \cite{hoffman:01}.

\subsection{Evaluation criteria}
The evaluation was based on the classical metrics of quality and time used in IPC. Time score is computed as the quotient $T*/T$ where $T*$ is the minimum time required by the planner to solve the problem, and $T$ is the time spent by the evaluated implementation to solve the same planning task. Quality score is computed as $Q*/Q$ where $Q*$ is the cost of the best known plan for a particular problem and Q is the cost of the plan produced by the evaluated implementation. If the planner found no solution the quality is set to zero.

\subsection{Results}
\begin {figure}[!h]
\centering
\includegraphics[scale=0.35]{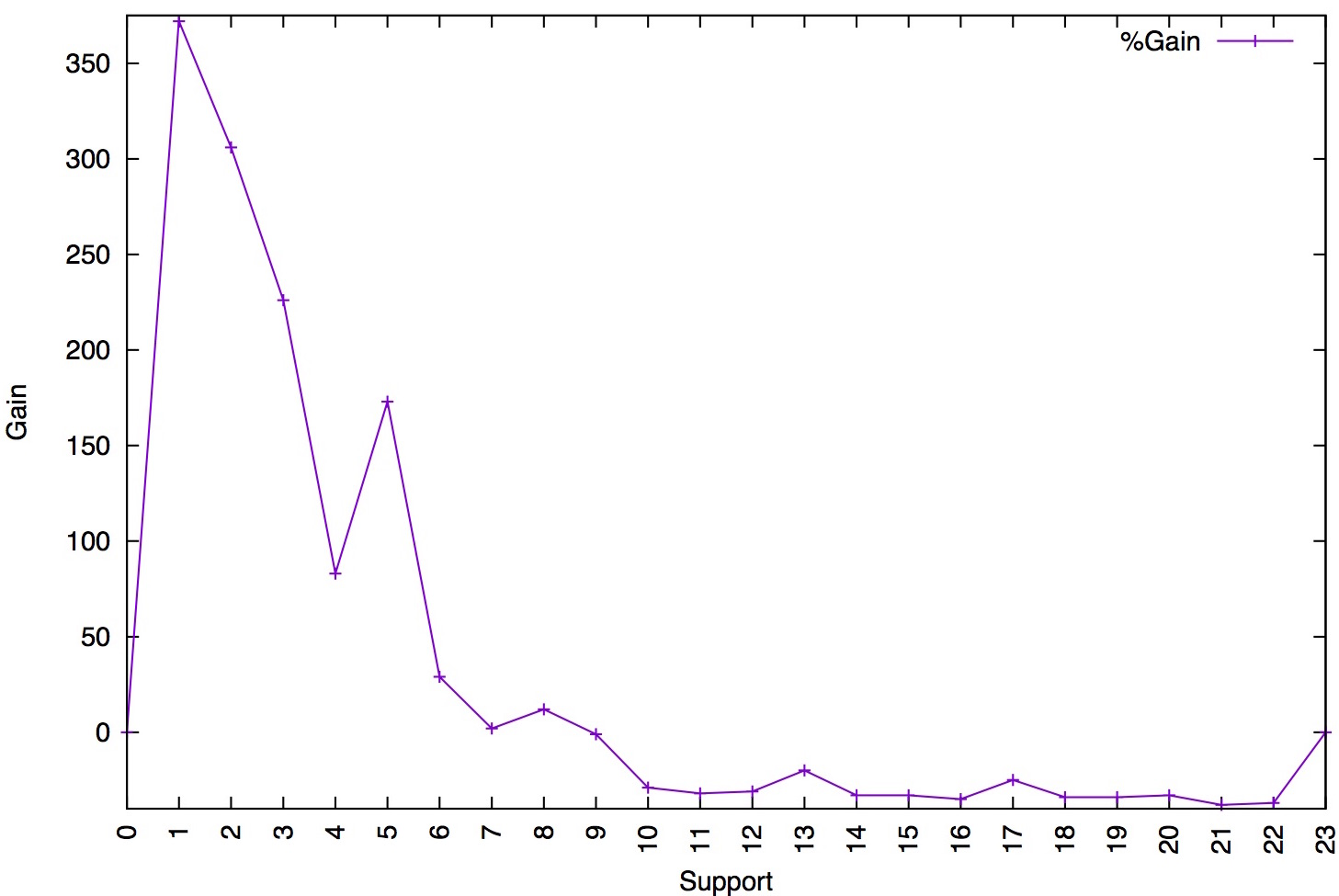}
\vspace{-1em}
\caption {Barman domain.}
\vspace{-1em}
\label{Fig:barman}
\end {figure}

\begin {figure}[!h]
\centering
\includegraphics[scale=0.35]{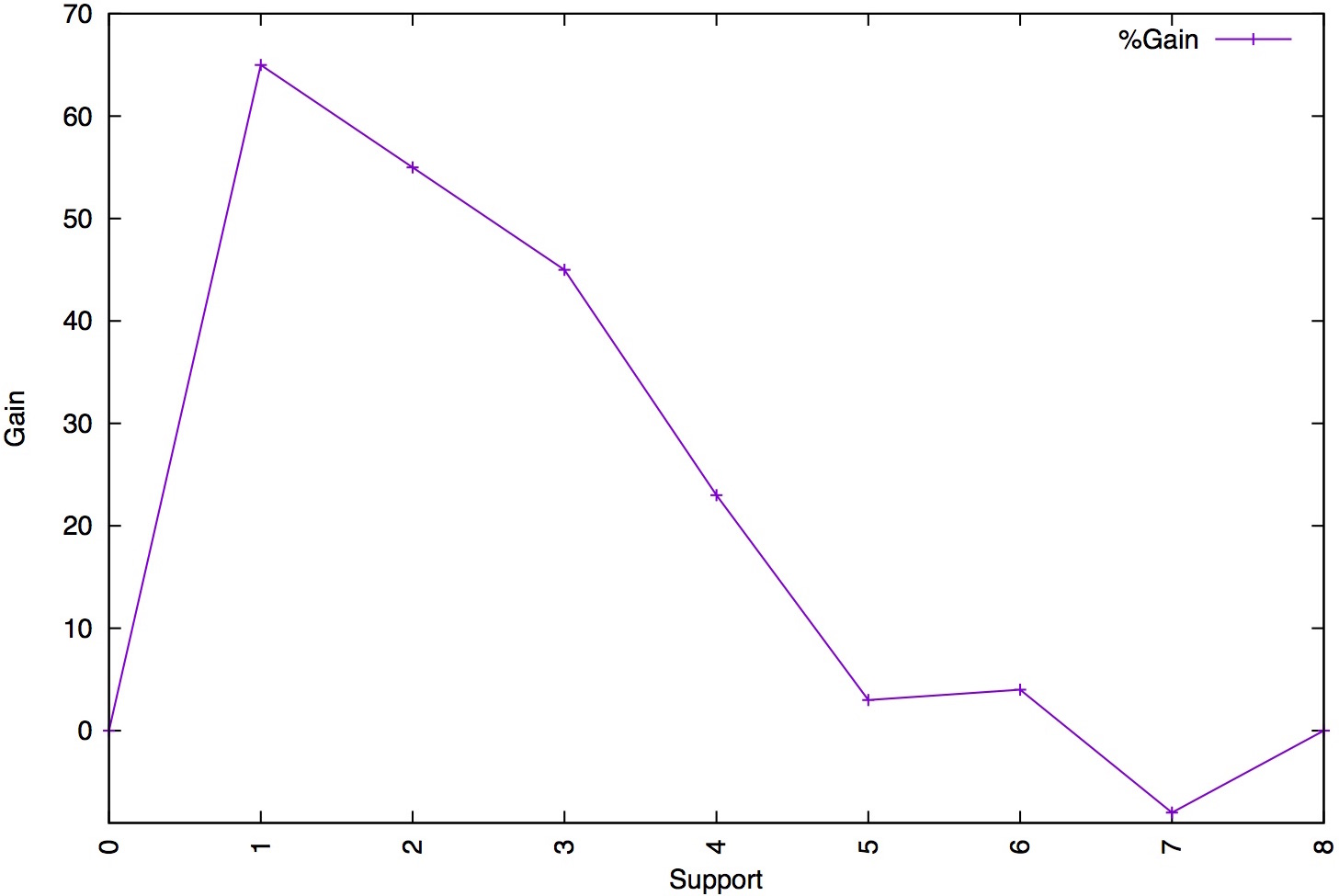}
\vspace{-1em}
\caption {Blocksworld domain.}
\vspace{-1em}
\label{Fig:blocksworld}
\end {figure}

\begin {figure}[!h]
\centering
\includegraphics[scale=0.35]{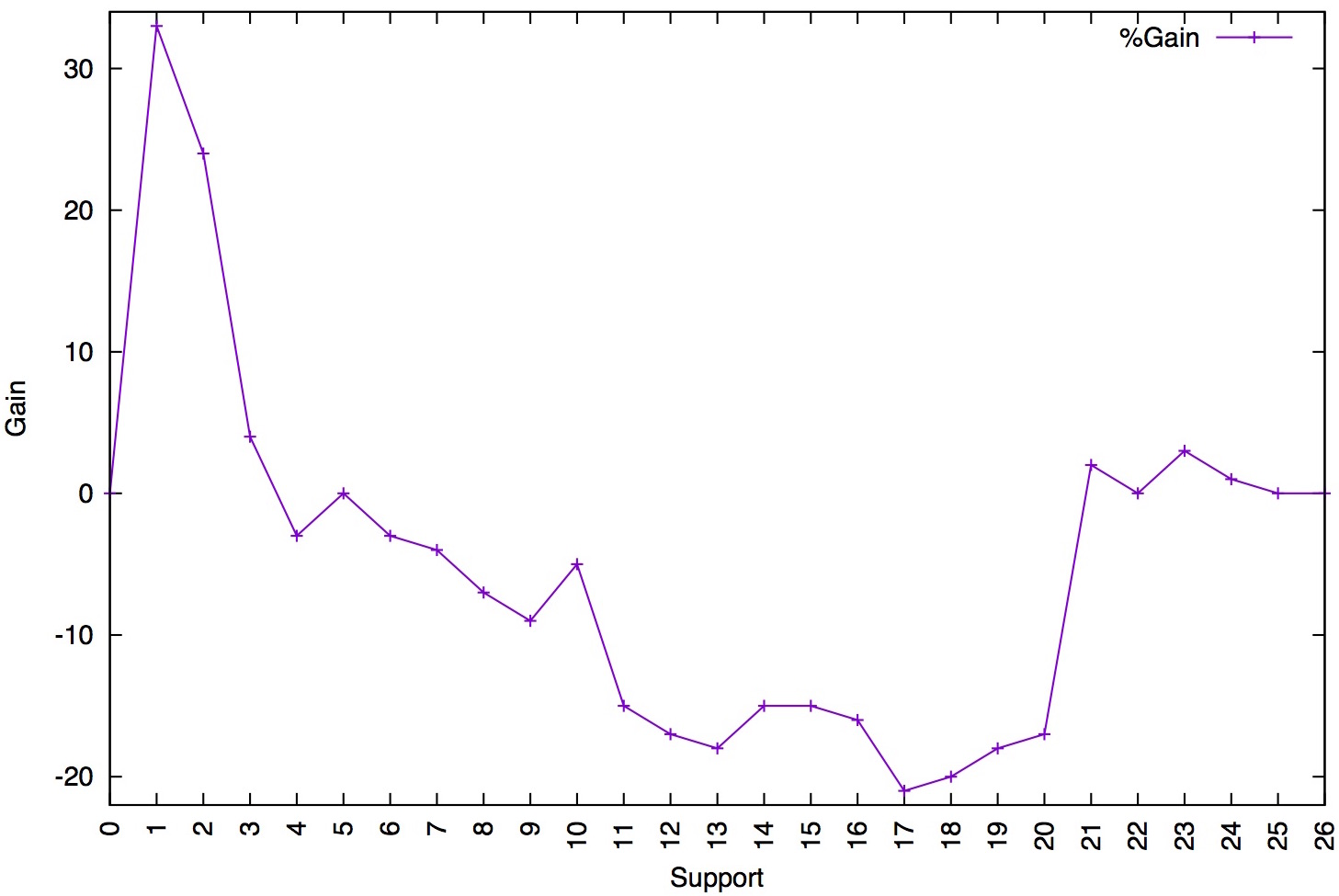}
\vspace{-1em}
\caption {Depots domain.}
\vspace{-1em}
\label{Fig:depots}
\end {figure}

\begin {figure}[!h]
\centering
\includegraphics[scale=0.35]{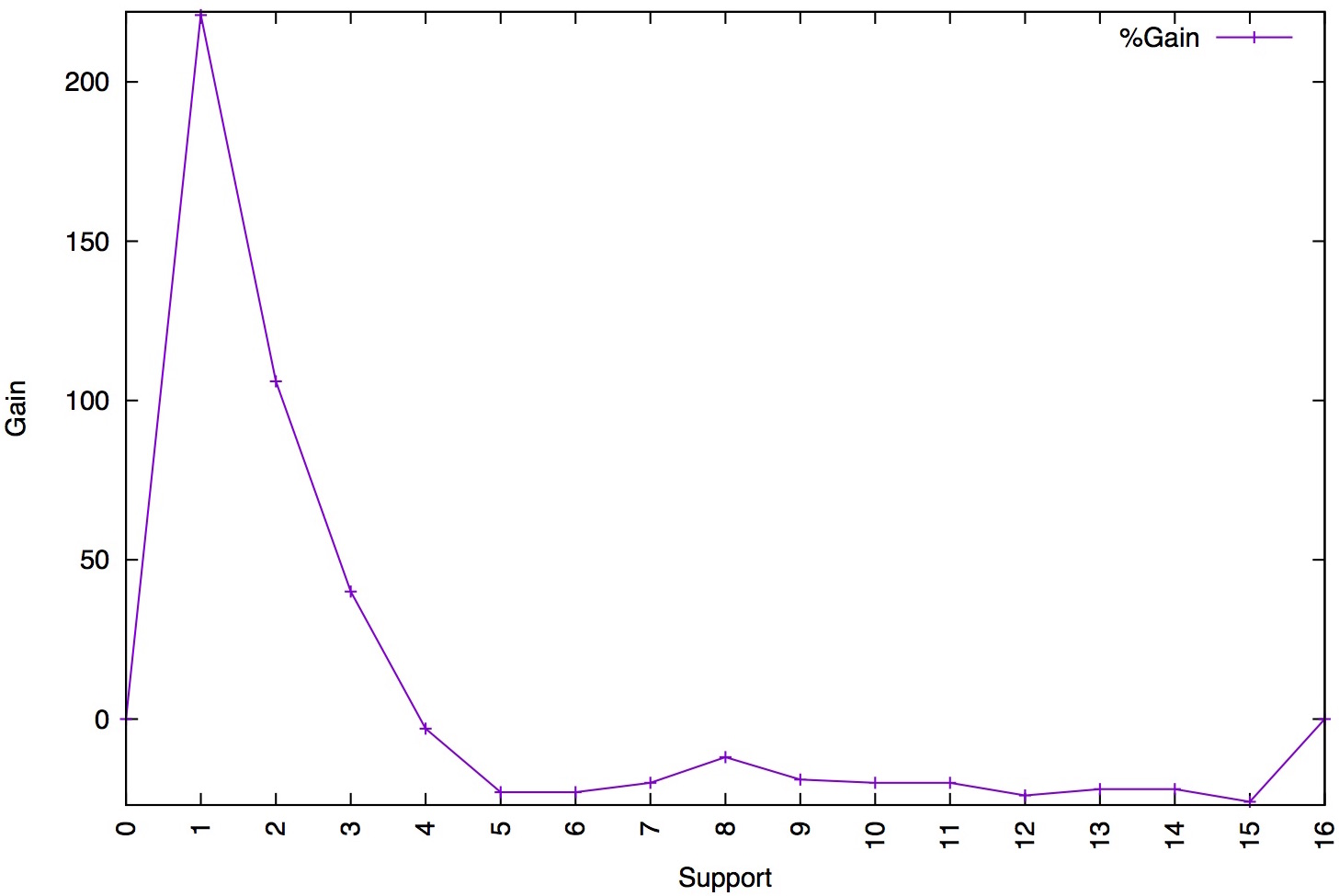}
\vspace{-1em}
\caption {Ferry domain.}
\vspace{-1em}
\label{Fig:ferry}
\end {figure}

\begin {figure}[!h]
\centering
\includegraphics[scale=0.35]{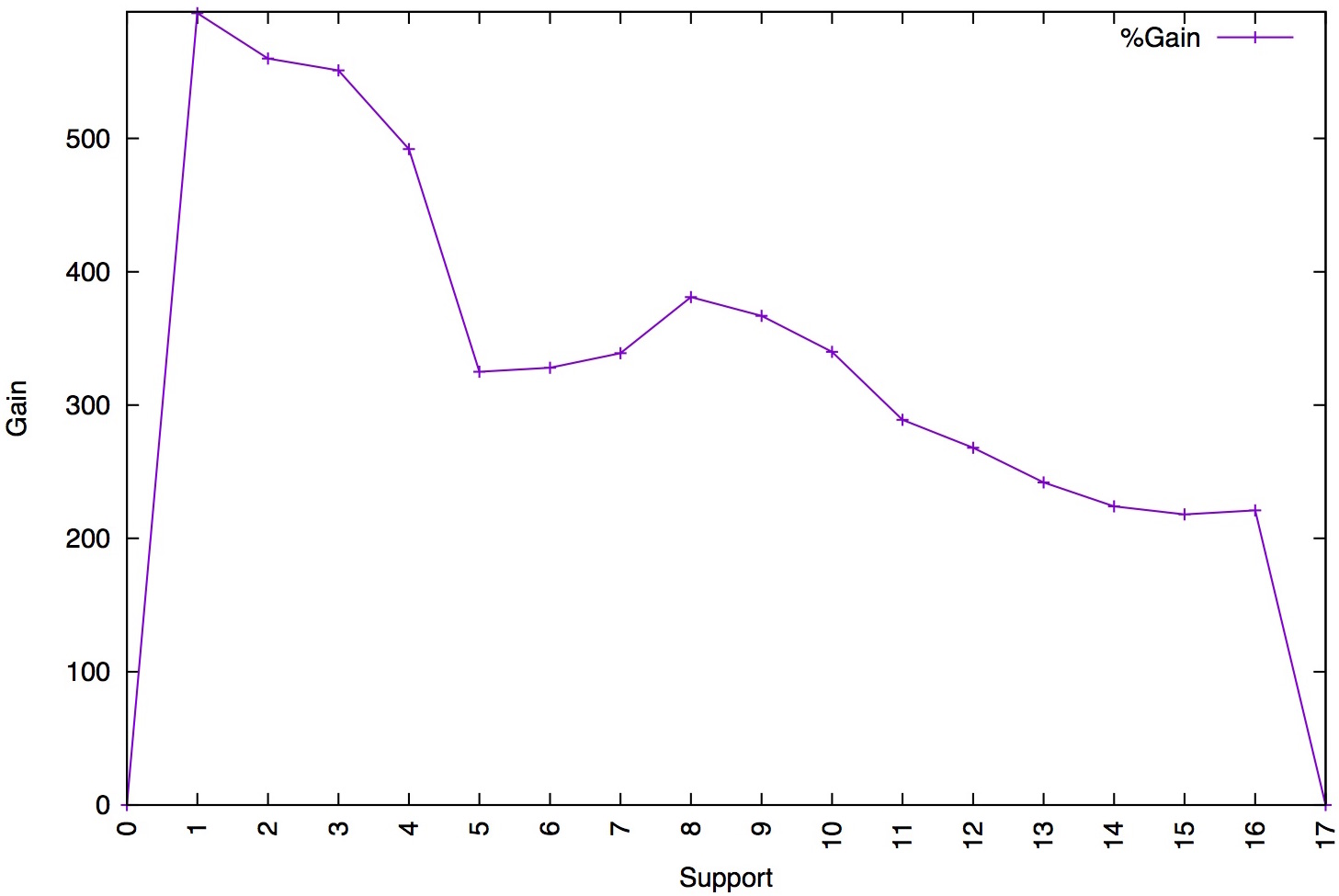}
\vspace{-1em}
\caption {Grid domain.}
\vspace{-1em}
\label{Fig:grid}
\end {figure}

\begin {figure}[!h]
\centering
\includegraphics[scale=0.35]{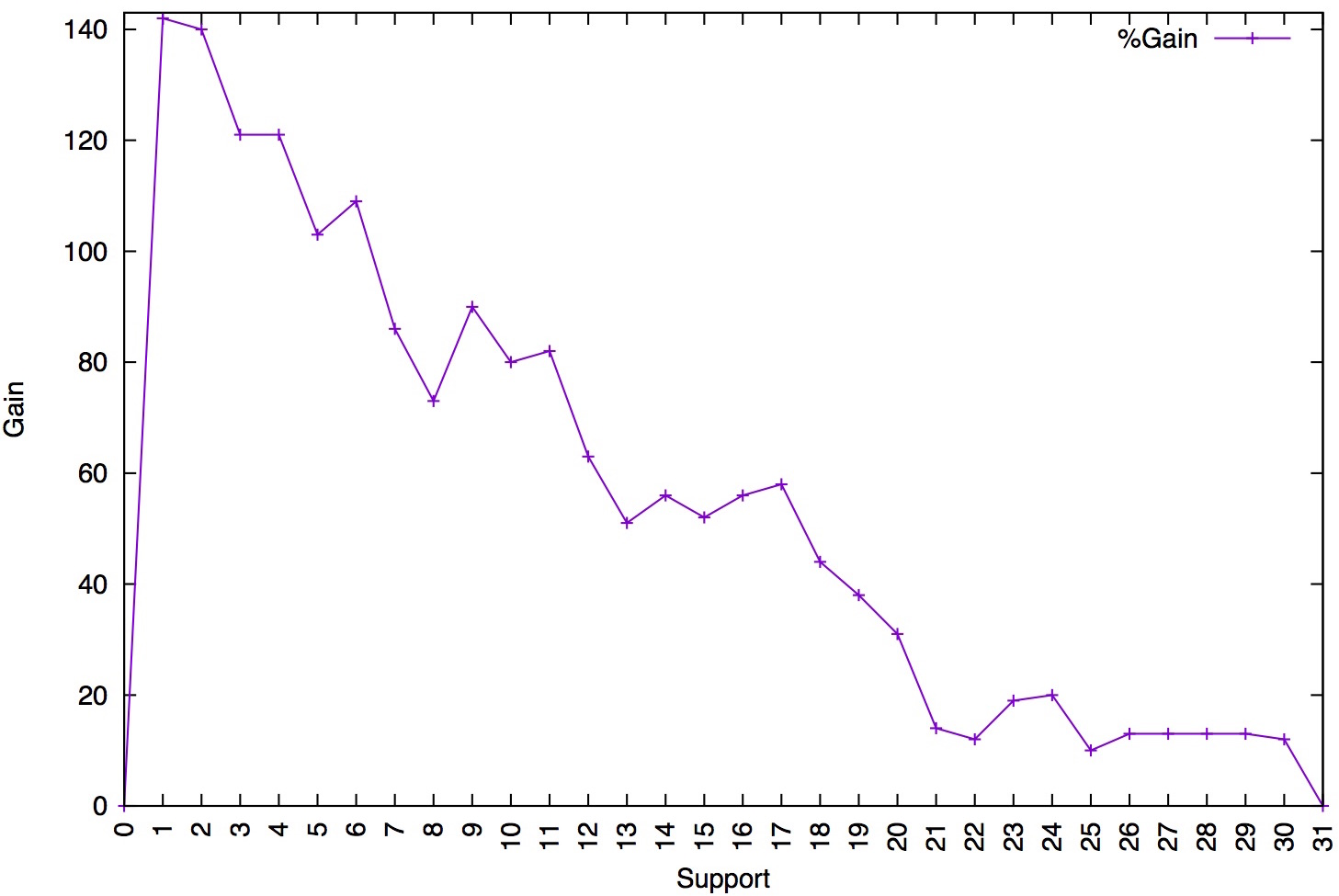}
\vspace{-1em}
\caption {Sokoban domain.}
\vspace{-1em}
\label{Fig:sokoban}
\end {figure}

Figures \ref{Fig:barman}, \ref{Fig:blocksworld}, \ref{Fig:depots}, \ref{Fig:ferry}, \ref{Fig:grid} and \ref{Fig:sokoban} report the gain of our method compared with the original algorithm among different support values. We can observe that large gains are obtained using lower values of the support parameter. Table\ref{Table:Evaluation} presents the results as relative gains with respect to the original algorithm given a support of 1\%.

\hfill \break

\begin{table}[!h]
\centering
\begin{tabular}{l r r r r r r r r}
 \hline
 Domain & Time & Quality \\ [0.5ex]
 \hline\hline
 Barman & 372\% & 78\%\\
 Blocksworld & 65\% & -4\%\\
 Depots & 34\%& 4\%\\
 Ferry & 221\%&-6\%\\
 Grid & 595\% & 0\%\\
 Sokoban & 142\%&-12\%\\ [1ex]
 \hline
\end{tabular}
\vspace{1em}
\caption{Improvement with 1\% support}
\vspace{-2em}
\label{Table:Evaluation}
\end{table}

\subsection{Discussion}
\begin{sloppypar}
Macro-actions quality score was improved in some domains owing the fact that the original algorithm could not found a plan for some problems (barman, blocksworld and depots). However, a domain where all the problems were solved by the original algorithm shows the worst quality (sokoban). It is due to the utility problem, i.e., the increasing of the branching factor due to the adding of macro-actions.

In our results, time improvement is obtained with a support of 1\% for all the domains. The results suggest that: (1) it can be possible to identify potential macro-actions over a domain by fixing the degree of support between 1\% and 30\%; (2) search performance can be improved, thus validating the relevance of macro-actions learning in the planning search. (3) the use of pattern mining algorithms is a good strategy to efficiently identify macro-actions.

Further work includes:
\begin{itemize}
\item The formalization of useful macro-actions as macro-operators based on the generalization of frequent sequences.
\item Implementing in our approach a way to deal with the utility problem. One promising direction is to apply outer entanglements \cite{Chrpa:2015} on macro-actions to reduce the number of instances of macros exploited.
\item The development of a solution to learn on-line macro-actions.
\end{itemize}

\end{sloppypar}

%
%
\section{Conclusion}
We proposed a novel planner independent macro learning method. It starts by automatically detecting useful macro-actions trough sequential pattern mining algorithms. We give special attention to the VMSP algorithm which finds maximal sequences as optimal sets of ordered sequences of actions.
It then implements those macro-actions in a forward chaining planner based on A* algorithm and on FF heuristic.

We have demonstrated that our method can be used to speed-up the search with a solid, and sometimes dramatic, gain compared to not using macro-actions. This stands for the six studied domains, specially Grid, Barman, Ferry and Sokoban.

%
%

\end{document}